\documentclass[]{mosi}

\usepackage[T1]{fontenc}
\usepackage{mathpazo} 

\usepackage{amsmath} 
\usepackage{natbib}
\usepackage{graphicx}
\usepackage{subcaption} 
\usepackage{caption}

\usepackage[toc,page,header]{appendix}
\usepackage[utf8]{inputenc} 
\usepackage[T1]{fontenc}    
\usepackage{hyperref}       
\usepackage{url}            
\usepackage{booktabs}       
\usepackage{amsfonts}       
\usepackage{nicefrac}       
\usepackage{microtype}      
\usepackage{wrapfig}

\usepackage{amssymb}        
\usepackage{titletoc}
\usepackage{minitoc}

\usepackage{array}
\usepackage{etoolbox}

\definecolor{lightblue}{RGB}{200, 230, 255}  
\definecolor{headerblue}{RGB}{150, 200, 255} 

\usepackage{pgfplots}
\usepackage{xcolor}
\usepackage{float}
\usepackage{comment}
\usepackage{multirow}
\usepackage{makecell}
\usepackage{siunitx}
\usepackage{tikz}
\usepackage{pgf-pie}
\usepackage[export]{adjustbox}

\usepackage{ragged2e}
\usepackage{tabularx}
\usepackage{caption}
\usepackage{enumitem}
\usepackage{pifont}
\usepackage[hang,flushmargin]{footmisc}

\usepackage{tcolorbox}
\tcbuselibrary{breakable}
\tcbuselibrary{skins}

\usepackage{tabularx}
\usepackage{listings}
\usepackage{arydshln}
\usepackage{algorithm}
\usepackage[noend]{algpseudocode}

\makeatletter
\def\BState{\State\hskip-\ALG@thistlm}
\makeatother

\makeatletter
\def\adl@drawiv#1#2#3{%
        \hskip.5\tabcolsep
        \xleaders#3{#2.5\@tempdimb #1{1}#2.5\@tempdimb}%
                #2\z@ plus1fil minus1fil\relax
        \hskip.5\tabcolsep}
\newcommand{\cdashlinelr}[1]{%
  \noalign{\vskip\aboverulesep
           \global\let\@dashdrawstore\adl@draw
           \global\let\adl@draw\adl@drawiv}
  \cdashline{#1}
  \noalign{\global\let\adl@draw\@dashdrawstore
           \vskip\belowrulesep}}
\makeatother

\usepackage{threeparttable}
\usepackage{setspace}

\definecolor{MossCyan}{HTML}{82D9FF} 
\definecolor{MossBlue}{HTML}{82B1FF}

\definecolor{ForestGreen}{RGB}{34, 139, 34}
\definecolor{Red}{RGB}{255, 0, 0}

\definecolor{tickG}{rgb}{0.1, 0.588, 0.1}
\definecolor{crossR}{rgb}{0.588, 0.1, 0.1}

\definecolor{frenchblue}{rgb}{0.0, 0.45, 0.73}
\definecolor{babyblue}{rgb}{0.54, 0.81, 0.94}
\definecolor{classicrose}{rgb}{0.98, 0.8, 0.91}
\definecolor{beige}{rgb}{0.96, 0.96, 0.86}
\definecolor{forestgreen}{HTML}{2e7d43}

\definecolor{blue1}{HTML}{91BBE6}
\definecolor{blue2}{HTML}{3F90E0}
\definecolor{blue3}{HTML}{316FAD}

\definecolor{color1}{HTML}{FF9999}
\definecolor{color2}{HTML}{FF6666}
\definecolor{color3}{HTML}{FF3333}
\definecolor{color4}{HTML}{E60000}
\definecolor{color5}{HTML}{B30000}
\definecolor{color6}{HTML}{8CD98C}
\definecolor{color7}{HTML}{53c653}
\definecolor{color8}{HTML}{00B050}
\definecolor{color9}{HTML}{2d862d}
\definecolor{color10}{HTML}{206020}
\definecolor{color11}{HTML}{cca300}

\newcommand{\cplus}[1]{\textcolor{green!60!black}{\footnotesize +#1}}
\newcommand{\cminus}[1]{\textcolor{red}{\footnotesize -#1}}
\newcommand{\cgray}[1]{\textcolor{gray}{\footnotesize #1}}

\newtcolorbox{promptbox}[2][]{
    colback=white,
    coltext=black,
    arc=3mm,
    boxrule=0.5pt,
    colframe=black!60!white,
    title={#2},
    colbacktitle=black,
    coltitle=white,
    fonttitle=\bfseries,
    top=8pt,
    bottom=8pt,
    left=10pt,
    right=10pt,
    breakable,
    before upper={%
        \linespread{1}\selectfont
        \setlength{\parskip}{1ex plus 0.2ex minus 0.2ex}%
        \setlength{\parindent}{0pt}%
    },
    #1
}

\title{FourierSampler: Unlocking Non-Autoregressive Potential in Diffusion Language Models via Frequency-Guided Generation}

\author{
Siyang He$^{1,2,3,*}$ \hspace{.3em}
Qiqi Wang$^{1,2,3,*}$\hspace{.3em}
Xiaoran Liu$^{1,2,3}$\hspace{.3em}
Hongnan Ma$^{3}$ \hspace{.1em}
\\
\textbf{
Yiwei Shi$^{3}$ \hspace{.1em}
Yuerong Song$^{1,2,3}$ \hspace{.1em}
Ying Zhu$^{1,2,3}$ \hspace{.1em}
Tianyi Liang$^{2,3}$ \hspace{.1em}
}
\\
\textbf{
Zengfeng Huang$^{1,2}$ \hspace{.2em}
Ziwei He$^{1,2,3,\dagger}$\hspace{.2em}
Xipeng Qiu$^{1,2,3,\dagger}$\hspace{.2em}
}
\\
[1ex]
$^{1}$Fudan University   
$^{2}$Shanghai Innovation Institute 
$^{3}$OpenMOSS Team 
\\
}

\abstract{
\begin{abstract}
'Despite the non-autoregressive potential of diffusion language models (dLLMs), existing decoding strategies demonstrate positional bias, failing to fully unlock the potential of arbitrary generation. In this work, we delve into the inherent spectral characteristics of dLLMs and present the first frequency-domain analysis showing that low-frequency components in hidden states primarily encode global structural information and long-range dependencies, while high-frequency components are responsible for characterizing local details. Based on this observation, we propose FourierSampler, which leverages a frequency-domain sliding window mechanism to dynamically guide the model to achieve a ``structure-to-detail" generation. FourierSampler outperforms other inference enhancement strategies on LLADA and SDAR, achieving relative improvements of 20.4\% on LLaDA1.5-8B and 16.0\% on LLaDA-8B-Instruct. It notably surpasses similarly sized autoregressive models like Llama3.1-8B-Instruct.
\end{abstract}
}

\checkdata[Correspondence]{\url{hsy2678501@gmail.com}, \url{qqwang7799@gmail.com}, \url{ziwei.he@sii.edu.cn}, \url{xpqiu@fudan.edu.cn}}
\checkdata[Repository]{\url{https://github.com/rimory6/FourierSampler}}

\begin{document}

\maketitle
\begingroup
\renewcommand{\thefootnote}{\fnsymbol{footnote}}
\setcounter{footnote}{1}
\footnotetext{Equal contribution.}
\endgroup




\section{Introduction}\label{sec:intro}

Diffusion Large Language Models (dLLMs)~\citep{sahoo2024simple,ou2024your,shi2024simplified} have emerged as a notable non-autoregressive paradigm. By breaking the strict left-to-right decoding constraint of traditional autoregressive (AR) models~\citep{gpt4,cai2024internlm2,dubey2024llama}, dLLMs utilize an arbitrary-order generation strategy. This flexibility allows the model to leverage global bidirectional context during the entire generation process, rather than being limited to prefix history. Consequently, this paradigm has demonstrated distinct advantages over AR models, including mitigating the ``reversal curse~\citep{berglund2023reversal}" inherent to unidirectional causal masking, facilitating controllable generation tasks such as text infilling, and enabling non-sequential planning for complex problem-solving.

\begin{wrapfigure}{r}{0.5\textwidth} 
    \centering
    \vspace{-15pt}
    \includegraphics[width=\linewidth]{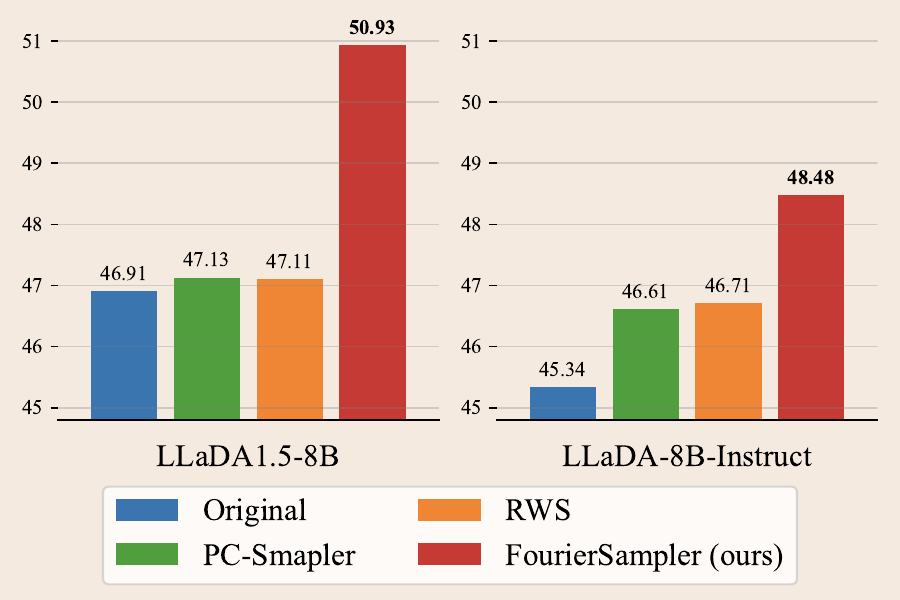}
    \vspace{-5pt}
    \caption{The average score in different tasks of FourierSampler compared with other decoding strategies in LLaDA~\citep{nie2025large,zhu2025llada}.}
    \label{fig:intro}
    \vspace{-15pt}
\end{wrapfigure}

Despite these benefits, existing research indicates that dLLMs often exhibit a strong positional bias. In fact, ~\citep{huang2025pc} have shown that constraining the model to left-to-right order can even surpass standard confidence-based decoding on several math tasks. Although empirically effective on specific benchmarks, such strict generation constraints the inherent flexibility of dLLMs, preventing them from fully leveraging global bidirectional context for holistic planning. To address this problem, recent works have adopted ad-hoc measures, including rule-based methods~\citep{huang2025pc} or reward-guided strategies~\citep{gwak2025reward}. Notably, the effectiveness of these heuristic approaches indicates that optimizing the decoding schedule is crucial to further enhance dLLM capabilities.


Different from the above works that calibrate the sampling distribution with external priors, in this work, we introduce a principled approach to mitigate the dominance of positional bias and unlock the intrinsic non-autoregressive capabilities of dLLMs. Diverging from reliance on external intervention signals, we delve into the frequency-domain characteristics of the model’s internal representations. Drawing on insights from the Fourier Transformer\cite{he2023fourier}, we observe that low-frequency components in textual representations typically encapsulate global structural information and long-range dependencies, whereas high-frequency components are responsible for characterizing local details. Based on this intuition, we propose FourierSampler, a decoding strategy that dynamically focuses on different frequency bands at different decoding steps, which guides dLLMs to first decode structural content dominated by low-frequency signals and subsequently complement detailed content dominated by high-frequency signals. This produces a beneficial long-range influence on the generation trajectory, thereby enhancing the overall quality and logical coherence of the output. As shown in Figure~\ref{fig:intro}, our method can statistically outperform the original model and other work relying on external signal-guided decoding on various math and code tasks. Our contribution can be summarized as follows.

\begin{itemize}
\item We conduct the first frequency analysis in dLLMs that the low-frequency components of hidden states in the temporal dimension correspond to structural information in the output, while the high-frequency components correspond to detailed information, which provides internal guidance for dLLM decoding.
\item We propose \textbf{FourierSampler}, our dLLM decoding sampling scheme, which guides the model to achieve a structure-to-detail decoding with the Translated Fourier Score, and balances the guidance with the original confidence by an Adaptive Fourier Calibrator.
\item We conduct validation experiments on two types of dLLMs, LLaDA and SDAR, and find that FourierSampler consistently achieves stable improvements in code and math tasks, specifically achieving relative improvements up to 20.4\% and 16.0\% on LLaDA1.5-8B and LLaDA-8B-Instruct, and up to 45.1\% and 26.5\% on SDAR-1.7B-Chat and SDAR-4B-Chat, respectively. These results surpass other inference enhancement
strategies as well as autoregressive models with similar sizes, providing new insights for an in-depth understanding of dLLM decoding enhancement.
\end{itemize}

\begin{figure*}[!t]
    \centering
    \includegraphics[width=0.9\linewidth]{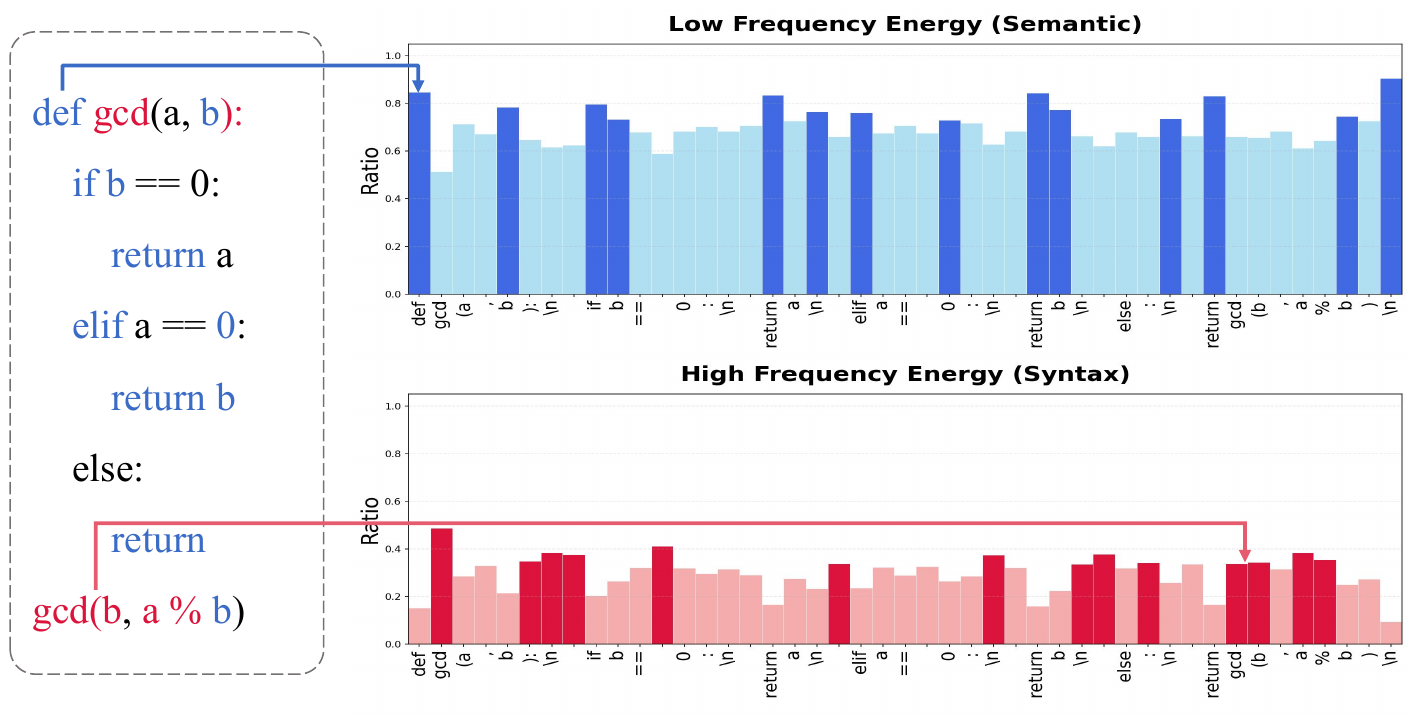}
    \caption{Visualization of the correspondence between frequency-domain analysis and textual information. In the hidden states after a forward pass, tokens dominated by low-frequency signals correspond to structural information like \textit{if} and \textit{elif}, while tokens dominated by high-frequency signals correspond to detailed information like \textit{gcd}.} 
    \label{fig:observation}
\end{figure*}

\section{Related Work}\label{sec:related}

\subsection{Decoding Strategy for dLLMs}

Diffusion Large Language Models (dLLMs)~\citep{sahoo2024simple,ou2024your,shi2024simplified} have become a hot topic in NLP. Models like LLaDA~\citep{nie2025large}, Dream~\citep{ye2025dream}, Mercury~\citep{mercury}, Gemini Diffusion~\citep{genimid} and SDAR~\citep{cheng2025sdar} confirm the scalability of this paradigm~\citep{nie2024scaling,gong2024scaling,ni2025training} and fuel intensive follow-up work on long-context modeling~\citep{liu2025longllada,he2025ultrallada}, inference efficiency~\citep{wu2025fast,wu2025fast2,song2025sparse}, multimodal extensions~\citep{you2025llada,yang2025mmada}, and post-training strategies~\citep{zhu2025llada,zhao2025d1,wang2025revolutionizing,zhu2025dirl}. Although dLLMs can decode in any order, their decoding trajectories still exhibit pronounced positional bias~\citep{gwak2025reward}.
\noindent 
Consequently, recent work focuses on optimizing token unmasking orders to enhance generation planning. Mainstream models like LLaDA~\citep{nie2025large} and SDAR~\citep{cheng2025sdar} adopt confidence-based unmasking, with variants prioritizing unmasking strategies based on maximum probability, entropy~\citep{ben2025accelerated}, or confidence gaps~\citep{kim2025train}, alongside random sampling baselines~\citep{austin2021structured}.

To enhance generation performance, advanced methods introduce external interventions. For example, PC-Sampler~\citep{huang2025pc} uses rules-based biases for specific positions, while RWS~\citep{gwak2025reward} employs reward models to enhance coherence. Alternatively, training-based approaches like DOT~\citep{ye2024diffusion}, DDPD~\citep{liu2024think}, and DCoLT~\citep{huang2025reinforcing} optimize generation trajectories via post-training or reinforcement learning. However, these methods rely on complex external signals or costly training. They overlook the potential of mining dLLM internal representations for effective decoding guidance.

\subsection{Spectral Foundations of Transformers}

The representation power of Transformers can be deeply understood through the lens of signal processing, particularly by analyzing their behavior in the frequency domain. While FNet~\citep{lee2022fnet} and its successors~\citep{zhuang2022long, scribano2023dct} demonstrate the efficacy of spectral token-mixing, theoretical evidence suggests that multi-head self-attention inherently functions as a low-pass filter~\citep{wang2022anti, park2022vision}. This is further corroborated by Fourier Transformer~\citep{he2023fourier}, which observes that power spectra concentrate in low-frequency bins within deeper layers. However, while these spectral properties have been integrated into autoregressive architectures, their potential to guide the decoding process in dLLMs remains unexplored.

\section{Method}\label{sec:method}
\subsection{Spectral Semantic Analysis in dLLMs}\label{subsec:freq}
\noindent Building upon the theoretical foundations of spectral analysis (Sec \ref{sec:related}), we investigate how spectral properties manifest within the unique architecture of dLLMs. Given that low-frequency components typically carry global trends while high-frequency components encode fine-grained details, we hypothesize that the hidden states in dLLMs exhibit a similar semantic stratification across their frequency spectrum.

To validate this, we select two text samples possessing distinct structural features for analysis, a Python script for calculating the Greatest Common Divisor shown in Figure~\ref{fig:observation}, and a mathematical derivation of the difference of squares formula in Figure~\ref{fig:observation2}. After a single forward in dLLMs, we extract the final-layer hidden states $\bm{H} \in \mathbb{R}^{L \times D}$ where $L$ is the sequence length and $D$ is the hidden dimension, and compute the low-frequency component $\bm{H}'_\text{low}$ by retaining only the lowest-frequency half of the spectrum via a 
binary frequency mask $\bm{M} \in \{0,1\}^{W}$($W$ denotes the number of frequency components):
\begin{equation}
\begin{array}{ll}
\bm{H}'_\text{low} 
= \mathcal{F}_r^{-1}\!\left(\mathcal{F}_r(\bm{H}) \odot \bm{M}\right),
&
\bm{M}_k =
\begin{cases}
1, & 0 \le k < \lfloor W/2 \rfloor, \\
0, & \text{otherwise}.
\end{cases}
\end{array}
\label{fourierfilter}
\end{equation}

where $\mathcal{F}_r$ denotes the real-valued Fourier transform, since $\bm{H}$ is real-valued~\citep{sorensen2003real}. We then define the low-frequency ratio $r_\text{low}$ as:
\begin{equation}
\label{r_low}
r_\text{low} = \frac{\|\bm{H}'_\text{low}\|_2^2}{\|\bm{H}\|_2^2},
\end{equation}
where $\|\cdot\|_2$ denotes the Euclidean norm along the feature dimension $D$. Similarly, we also define the high-frequency ratio $r_\text{high} = 1 - r_\text{low} $, which accounts for the remaining spectral energy.

Based on this metric, we highlight the Top-14 tokens with the highest low-frequency and high-frequency ratios in Figure~\ref{fig:observation} and Figure~\ref{fig:observation2}. Observations indicate that in the code task, reserved keywords constituting the logical skeleton of the program (\textit{if}, \textit{elif}, \textit{return}) exhibit energy significantly concentrated in the low-frequency band. Conversely, specific function names and numerical values display distinct high-frequency characteristics. Similarly, in the math task, natural language text guiding the derivation logic is dominated by low-frequency signals, whereas specific mathematical formulas account for the majority of the high-frequency energy. 

This analysis confirms, for the first time in dLLMs, a semantic correspondence where low-frequency components capture the structural skeleton, while high-frequency components encode local details. This discovery implies that dLLM decoding can be optimized as a hierarchical refinement process. Specifically, prioritizing the generation of low-frequency ``structural" tokens provides a stable global context that constrains the search space for subsequent high-frequency ``detail" tokens. Conversely, attempting to determine fine-grained details before the logical skeleton is established can lead to structural inconsistencies or error propagation, which is a common failure mode in non-autoregressive generation.
Notably, this structure-to-detail generation paradigm has been extensively explored and validated in diffusion-based text-to-image generation~\citep{Gao_2024}~\citep{Besnier2025HaltonSF}, where models are widely observed to first form low-frequency global layouts and semantic compositions before progressively refining high-frequency textures and fine details.

Inspired by this hierarchical dependency, we implement a dynamic spectral filtering method called FourierSampler. By employing a frequency-domain sliding window that gradually transitions the passband from low to high frequencies, we enforce a structure-to-detail generation order. This ensures that the model first converges on a robust global plan before refining the specific nuances of the output.

\begin{figure*}[!t]
    \centering
    \includegraphics[width=0.9\linewidth]{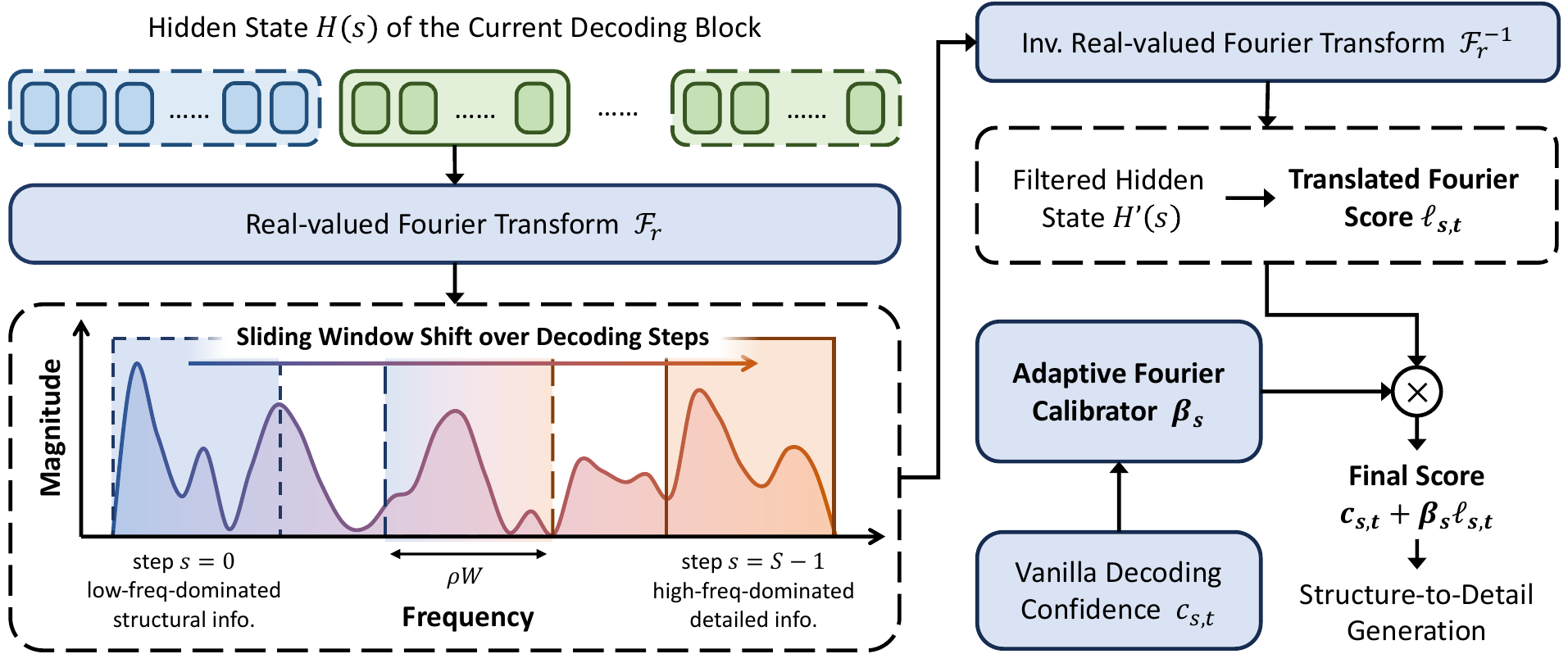}
    \caption{Overview of our FourierSampler. A sliding window in the frequency domain, retaining the low frequency at the beginning and the high frequency at the end based on the decoding step $s$, guides the dLLM to decode structural content first, then detailed content via Translated Fourier Score and Adaptive Fourier Calibrator.}
    \label{fig:main}
\end{figure*}

\subsection{Translated Filtering Score}

The Translated Filtering Score provides a frequency-domain criterion for ranking token positions during decoding.
At each decoding step, it retains a fixed-width frequency band whose location shifts uniformly from the low-frequency end toward the high-frequency end of the spectrum, and scores tokens according to their energy within the retained band.

Assume the decoding block size is $B$ and the number of decoding steps per block is $S$.
Let $\bm{H}(s)\in\mathbb{R}^{B\times D}$ denote the hidden states of the current block at decoding step $s$.
We apply a real-valued Fourier transform along the sequence dimension, filter the spectrum using a step-dependent binary mask $\bm{g}(s)$, and then transform back to the original space:
\begin{equation}
\bm{H}'(s)
=
\mathcal{F}_r^{-1}
\Big(
\mathcal{F}_r\big(\bm{H}(s)\big)
\odot
\bm{g}(s)
\Big)\text{,}
\end{equation}
where $\odot$ denotes element-wise multiplication.
The filtered hidden state $\bm{H}'(s)$ is the reconstructed time-domain representation after retaining only the frequency components selected at step $s$.

Let $W$ denote the number of frequency components.
We set the width of the frequency window as
\begin{equation}
w = \max\!\left(1,\;\lfloor \rho W \rfloor\right)\text{,}
\end{equation}
where $\rho \in (0,1]$ controls the relative bandwidth.

The window translates uniformly along the frequency axis over decoding steps.
Accordingly, the starting position of the window at decoding step $s$ is defined as
\begin{equation}
o_s = \left\lfloor \frac{s}{S-1}\,(W - w) \right\rfloor \text{.}
\end{equation}

The frequency mask $\bm{g}(s)\in\{0,1\}^{W}$ is then defined as
\begin{equation}
\bm{g}_p(s)
=
\begin{cases}
1, & o_s \le p < o_s + w, \\[0.5ex]
0, & \text{otherwise},
\end{cases}
\end{equation}
for frequency index $p \in \{0,\dots,W-1\}$.

As $s$ increases, the retained frequency band shifts monotonically from the beginning toward the end of the spectrum, resulting in a step-dependent spectral preference during decoding, favoring low-frequency-dominated tokens in the earlier stage and high-frequency-dominated tokens later.

For a token at position $t$ within the block ($0 \le t \le B-1$), we compute its energy under the filtered representation and normalize it within the block to obtain the Translated Filtering Score $\ell_{s,t}$, where $\epsilon$ is a small constant added for numerical stability:

\begin{equation}
\ell_{s,t} = \frac{\displaystyle \mathop{\scalebox{1.4}{$\Sigma$}}_{d=1}^D \left(\bm{H}'_{[t,d]}(s)\right)^2}{\displaystyle \max_{t'}\left( \mathop{\scalebox{1.4}{$\Sigma$}}_{d=1}^D \left(\bm{H}'_{[t',d]}(s)\right)^2 \right)+\epsilon}\text{.}
\end{equation}
The score $\ell_{s,t}$ measures the relative intensity of token $t$ under the frequency band selected at step $s$, and is used to guide the decoding priority within the block.

\subsection{Adaptive Fourier Calibrator}
To dynamically adjust the guidance strength of the translated filtering score $\ell_{s,t}$ in the sampler, we introduce an adaptive weight $\beta_s$ based on the original decoding confidence. At decoding step $s$, let the output prediction distribution be $p_{s,t}$. For the set of masked positions, $\mathcal{M}_s$, we take the maximum probability $q_{s,t}=\max_{v}{p_{s,t}(v)}$ in the prediction distribution at each position $t\in\mathcal{M}_s$, and treat its variance over $\mathcal{M}_s$ as the ability of dLLM to distinguish between the writing priorities of different positions in the current decoding state.
\begin{equation}
\sigma_s^2=\text{Var}\left(\left\{q_{s,t}\right\}_{t\in\mathcal{M}_s}\right)\text{.}
\end{equation}
We record the values of this method over the past 20 decoding steps, compute the percentile $P_s$ of the current variance within the history, and normalize it into $w_s\in (0,1)$ interval, the cumulative distribution function of the normal distribution. The process of normalization is detailed in Appendix~\ref{sec:more_details}, and, finally, the adaptive weight is defined as follows, where $\beta_{\min}$ and $\beta_{\max}$ are the minimum and maximum values of the adaptive weight, respectively.
\begin{equation}
\beta_s=\beta_{\min}+(1-w_s)\big(\beta_{\max}-\beta_{\min}\big)\text{.}
\end{equation}
Based on the above translated filtering score and adaptive weight calculation, we add it to the original confidence $c_{s,t}$ as step $s$ for token $t$ to obtain the fusion score, shown as follows. 
\begin{equation}
\tilde{c}_{s,t}=c_{s,t}+\beta_s\ell_{s,t}\text{.}
\end{equation}
This design ensures that when the confidence differences between different positions are large, the model's own decoding intention is relatively clear, and the frequency guidance automatically weakens. Conversely, the frequential prior is strengthened, thereby achieving an adaptive decoding scheduler.

\section{Experiment}\label{sec:exp}
\subsection{Setup}

\begin{table*}[!t]
\centering
\small
\tabcolsep=0.11cm
\begin{tabular}{l cr cr cr cr cr cr}
\toprule
 & \textbf{GSM8k} & & \textbf{Math} & & \textbf{MBPP} & & \textbf{HE} & & \textbf{CD} & &\textbf{ Avg.} &  \\ 
\midrule
\textbf{\textit{Llama3.1-8B-Instruct}} & 80.97 & & 41.60 & & 65.37 & & 54.27 & & 0.00 & & 48.44 &  \\ 
\textbf{\textit{Qwen2.5-7B-Instruct}} & 81.65 & & 49.20 & & 66.93 & & 52.44 & & 4.30 & & 50.90 &  \\ 
\midrule
\textbf{\textit{LLaDA1.5-8B}} & 79.83 & & 41.40 & & 42.02 & & 40.85 & & 30.47 & & 46.91 &  \\ 
+ PC-Sampler & 81.20 & \cplus{1.7\%} & 43.00 & \cplus{3.9\%} & \textbf{51.36} & \cplus{22.2\%} & 39.02 & \cminus{4.5\%} & 21.09 & \cminus{30.8\%} & 47.13 & \cplus{0.5\%} \\ 
+ RWS & 80.67 & \cplus{1.1\%} & 42.00 & \cplus{1.4\%} & 43.19 & \cplus{2.8\%} & 39.63 & \cminus{3.0\%} & 30.08 & \cminus{1.3\%} & 47.11 & \cplus{0.4\%} \\ 
+ FourierSampler (ours) & \textbf{81.80} & \cplus{2.5\%} & \textbf{44.20} & \cplus{6.8\%} & 50.58 & \cplus{20.4\%} & \textbf{43.29} & \cplus{6.0\%} & \textbf{34.77} & \cplus{14.1\%} & \textbf{50.93} & \cplus{8.6\%} \\ 
\midrule
\textbf{\textit{LLaDA-8B-Instruct}} & 78.24 & & 42.20 & & 41.25 & & 39.63 & & 25.39 & & 45.34 &  \\ 
+ PC-Sampler & 79.00 & \cplus{1.0\%} & 40.40 & \cminus{4.3\%} & \textbf{49.81} & \cplus{20.8\%} & 39.63 & \cgray{0.0\%} & 24.22 & \cminus{4.6\%} & 46.61 & \cplus{2.8\%} \\ 
+ RWS & \textbf{79.61} & \cplus{1.8\%} & 42.80 & \cplus{1.4\%} & 42.02 & \cplus{1.9\%} & 39.02 & \cminus{1.5\%} & \textbf{30.08} & \cplus{18.5\%} & 46.71 & \cplus{3.0\%} \\ 
+ FourierSampler (ours) & \textbf{79.61} & \cplus{1.8\%} & \textbf{45.20} & \cplus{6.8\%} & 47.86 & \cplus{16.0\%} & \textbf{40.85} & \cplus{3.1\%} & 28.90 & \cplus{13.8\%} & \textbf{48.48} & \cplus{7.1\%} \\
\bottomrule
\end{tabular}
\caption{Results on LLaDA Series including LLaDA1.5-8B~\citep{zhu2025llada} and LLaDA-8B-Instruct~\citep{nie2025large} with best values in bold and relative improvement over vanilla decoding based on confidence colored in green (positive) and red (negative). Notably, FourierSampler achieves consistent positive improvements across all evaluated tasks, resulting in the best average performance. It also surpasses other competitive decoding strategies and similarly sized autoregressive models, including Llama3.1-8B-Instruct~\citep{dubey2024llama} and Qwen2.5-7B-Instruct~\citep{qwen2024qwen25technicalreport}.}
\label{tab:main_llada}
\end{table*}

\begin{table*}[!t]
\centering
\small
\tabcolsep=0.11cm
\begin{tabular}{l cr cr cr cr cr cr}
\toprule
 & \textbf{GSM8k} & & \textbf{Math} & & \textbf{MBPP} & & \textbf{HE} & & \textbf{CD} & & \textbf{Avg.} &  \\
\midrule
\textbf{\textit{Llama3.2-3B-Instruct}} & 69.37 & & 36.60 & & 50.97 & & 26.83 & & 0.00 & & 36.75 &  \\
\textbf{\textit{Qwen2.5-3B-Instruct}} & 76.42 & & 39.20 & & 47.47 & & 32.93 & & 10.94 & & 41.39 &  \\
\midrule
\textbf{\textit{SDAR-4B-Chat}} & 86.58 & & 48.20 & & 42.02 & & 57.93 & & 13.28 & & 49.60 &  \\
+ RWS & 87.41 & \cplus{1.0\%} & 49.20 & \cplus{2.1\%} & 39.69 & \cminus{5.5\%} & 60.37 & \cplus{4.2\%} & 16.02 & \cplus{20.6\%} & 50.54 & \cplus{1.9\%} \\
+ FourierSampler (ours) & \textbf{87.64} & \cplus{1.2\%} & \textbf{50.00} & \cplus{3.7\%} & \textbf{47.47} & \cplus{13.0\%} & \textbf{62.20} & \cplus{7.4\%} & \textbf{16.80} & \cplus{26.5\%} & \textbf{52.82} & \cplus{6.5\%} \\
\midrule
\textbf{\textit{SDAR-1.7B-Chat}} & 72.93 & & 39.60 & & 35.41 & & 37.80 & & 15.62 & & 40.27 &  \\
+ RWS & \textbf{75.59} & \cplus{3.6\%} & \textbf{41.00} & \cplus{3.5\%} & 35.80 & \cplus{1.1\%} & 31.71 & \cminus{16.1\%} & 17.97 & \cplus{15.0\%} & 40.41 & \cplus{0.3\%} \\
+ FourierSampler (ours) & 73.84 & \cplus{1.2\%} & 40.00 & \cplus{1.0\%} & \textbf{36.58} & \cplus{3.3\%} & \textbf{43.29} & \cplus{14.5\%} & \textbf{22.66} & \cplus{45.1\%} & \textbf{43.27} & \cplus{7.4\%} \\
\bottomrule
\end{tabular}
\caption{Results on SDAR Series including SDAR-4B-Chat and SDAR-1.7B-Chat~\citep{zhu2025llada} with best values in bold and relative improvement over vanilla decoding based on confidence colored in green (positive) and red (negative). Notably, FourierSampler achieves consistent positive improvements across all evaluated tasks, resulting in the best average performance. It also surpasses other competitive decoding strategies and similarly sized autoregressive models, including Llama3.2-3B-Instruct~\citep{meta2024llama} and Qwen2.5-3B-Instruct~\citep{qwen2024qwen25technicalreport}.}
\label{tab:main_sdar}
\end{table*}

We conduct experiments on widely used diffusion-based dLLMs, including LLaDA1.5-8B~\citep{zhu2025llada} and LLaDA-8B-Instruct~\citep{nie2025large}. In addition, to demonstrate that our method applies to models with other dLLM architectures, namely dLLMs with block-wise causal attention, we also evaluate our FourierSampler on SDAR-4B-Chat and SDAR-1.7B-Chat~\citep{cheng2025sdar}. The evaluation benchmarks include GSM8K (4-shot)~\citep{cobbe2021gsm8k}, MATH (4-shot)~\citep{hendrycks2021math}, MBPP (3-shot)~\citep{austin2021program}, HumanEval (0-shot)~\citep{chen2021humaneval}, and Countdown(0-shot)~\citep{ye2024beyond,ye2025longproc}. During evaluation, we set the default block size to 64 for all dLLMs and adopt OpenCompass~\citep{2023opencompass} as the evaluation framework. We use 512 generation steps for GSM8K, MATH, HumanEval, and MBPP, and 128 generation steps for Countdown. All experiments are conducted on NVIDIA H200 GPUs.

For the LLaDA Series, we use PC-Sampler~\citep{huang2025pc} and RWS~\citep{gwak2025reward} as main baselines. For the SDAR Series, since PC-Sampler does not provide the token frequency distribution for SDAR, we use RWS as the primary baseline. The relevant coefficients in PC-Sampler follow the settings in its original paper. For RWS, we adopt GRM-Llama3.2-3B~\citep{yang2024regularizing} as its reward model, recommended by its paper. We also compare dLLMs enhanced via different decoding strategies with similarly sized autoregressive models from Llama and Qwen Series~\citep{dubey2024llama,meta2024llama,qwen2024qwen25technicalreport}.

\subsection{Main Results}

For LLaDA1.5-8B and LLaDA-8B-Instruct, which are dLLMs based on full bidirectional attention, the experimental results in Table~\ref{tab:main_llada} demonstrate that our method consistently and significantly outperforms the baseline across different math and code tasks. In particular, we observe substantial relative improvements of up to 7.2\% on MATH, 20.4\% on MBPP, and 14.1\% on Countdown compared to the baseline. Moreover, our approach achieves the highest average performance across all benchmarks, surpassing other competitive methods.

Notably, our FourierSampler further enables LLaDA1.5-8B to bridge and exceed the performance gap with similarly sized autoregressive models, such as Llama3.1-8B-Instruct and Qwen2.5-7B-Instruct, where LLaDA1.5-8B originally underperformed on average, and other decoding strategies fail to achieve it. This result highlights that our method can more effectively unlock the potential of non-autoregressive generation in dLLMs.

For dLLMs with block-wise causal attention, including SDAR-4B-Chat and SDAR-1.7B-Chat, our method also consistently outperforms the baseline across all evaluated benchmarks in Table~\ref{tab:main_sdar}. Specifically, we achieve relative improvements of 3.7\% on MATH, 14.5\% on HumanEval, and 45.1\% on Countdown. In addition, the average performance across tasks is superior to that of other competing approaches. These results further demonstrate that our method generalizes well across different dLLM designs, including both full-bidirectional-attention and block-wise causal attention.

\subsection{Ablation Study}\label{sec:ablation}

\begin{figure}[!t]
    \centering
    \small
    
    \begin{minipage}[t]{0.48\linewidth}
        \vspace{0pt} 
        \centering
        
        \tabcolsep=0.2cm
        \begin{tabular}{lccc}
            \toprule
             & GSM8k & MBPP & Avg. \\
            \midrule
            \textbf{\textit{LLaDA1.5-8B}} & 79.83 & 42.02 & 60.97 \\ 
            + FourierSampler & \textbf{81.80} & \textbf{50.58} & \textbf{66.19} \\ 
            \cdashlinelr{1-4}
            + Fixed $\beta=0.4$ & 81.12 & 49.81 & 65.47 \\ 
            + Fixed $\beta=0.5$ & 81.12 & 47.86 & 64.49 \\ 
            + Fixed $\beta=0.6$ & 81.20 & \underline{50.19} & \underline{65.70} \\ 
            \cdashlinelr{1-4}
            + $\rho=0.4$ & \underline{81.35} & 42.80 & 62.08 \\ 
            + $\rho=0.6$ & 81.05 & 43.00 & 62.03 \\ 
            \midrule
            \textbf{\textit{LLaDA-8B-Instruct}} & 78.24 & 41.25 & 59.75 \\ 
            + FourierSampler & \underline{79.61} & \textbf{47.86} & \textbf{63.74} \\ 
            \cdashlinelr{1-4}
            + Fixed $\beta=0.4$ & 78.85 & 44.75 & 61.80 \\ 
            + Fixed $\beta=0.5$ & 79.38 & 46.69 & 63.04 \\ 
            + Fixed $\beta=0.6$ & 79.23 & \underline{47.08} & \underline{63.16} \\ 
            \cdashlinelr{1-4}
            + $\rho=0.6$ & \textbf{80.36} & 45.91 & 63.14 \\ 
            \bottomrule
        \end{tabular}
        \captionof{table}{Results on LLaDA Series~\citep{nie2025large,zhu2025llada} for the ablation study of FourierSampler.}
        \label{tab:ablation_dllm}
        \vspace{0.4cm} 
        
        \tabcolsep=0.11cm
        \begin{tabular}{l c c r c c r}
            \toprule
            \multirow{2}{*}{\textbf{ }} & 
            \multicolumn{3}{c}{\textbf{GSM8K}} & 
            \multicolumn{3}{c}{\textbf{MBPP}} \\
            \cmidrule(lr){2-4}\cmidrule(lr){5-7}
            & \textbf{Vanilla} & \textbf{Ours} & $\Delta(\%)$ 
            & \textbf{Vanilla} & \textbf{Ours} & $\Delta(\%)$ \\
            \midrule
            $B=16$   & 80.82 & 81.05 & \footnotesize +0.3\% & 49.03 & 48.25 & \footnotesize -1.6\% \\
            $B=32$   & 80.06 & 80.36 & \footnotesize +0.3\% & 49.81 & 51.36 & \footnotesize +3.1\% \\
            $B=64$   & 79.83 & 81.80 & \footnotesize +2.5\% & 42.02 & 50.58 & \footnotesize +20.4\% \\
            $B=128$  & 68.16 & 72.55 & \footnotesize +9.5\% & 37.74 & 44.75 & \footnotesize +18.6\% \\
            \bottomrule
        \end{tabular}
        \captionof{table}{Results with different block sizes on LLaDA1.5.}
        \label{tab:blocksize_ablation}
    \end{minipage}
    \hfill 
    \begin{minipage}[t]{0.48\linewidth}
        \vspace{0pt} 
        \centering
        \includegraphics[width=1.0\linewidth, height=5cm]{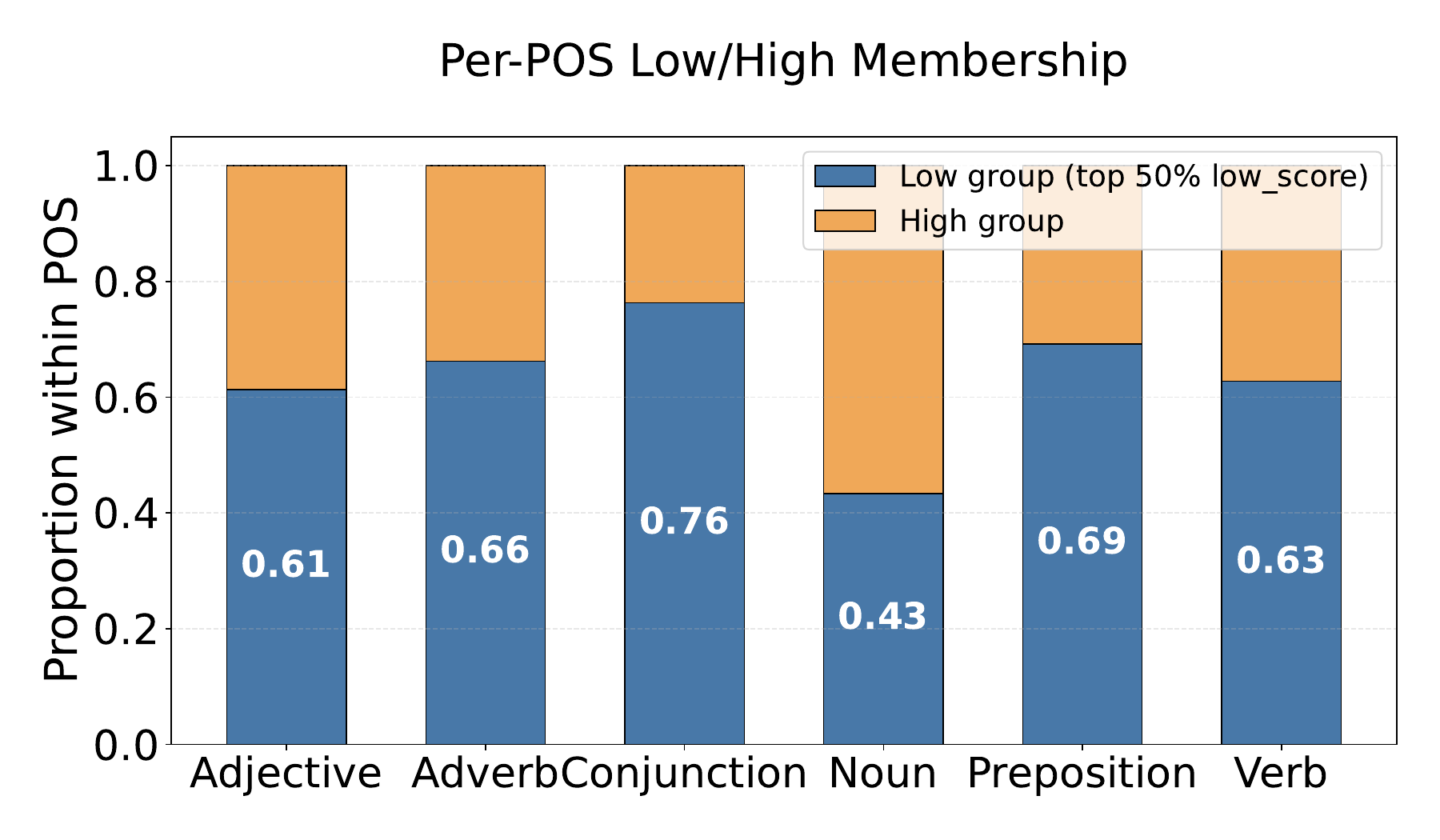}
        \caption{
            Low- and high-frequency features of different parts of speech in the frequency domain.
            The blue bars represent the proportion of tokens classified into the \textit{low group}, while the orange bars represent the \textit{high group}.
            Observations indicate that functional words responsible for syntactic structure, like conjunctions, exhibit dominant low-frequency features. In contrast, nouns, which typically serve as specific content fillers, show the strongest high-frequency tendency.
            This distribution corroborates our hypothesis that low-frequency components encode the structural skeleton, while high-frequency components correspond to detailed entities.
        }
        \label{fig:partofspeech}
    \end{minipage}
\end{figure}

To verify the rationality of the Adaptive Fourier Calibrator module design in our method, we conduct ablation studies by fixing the parameter $\beta$ to the maximum, minimum, and mean values of the adaptive weights, respectively, and evaluate the performance on GSM8K and MBPP. The results in Table~\ref{tab:ablation_dllm} show that different tasks on the same model may prefer different weight values. However, using fixed weights consistently underperforms the adaptive weighting strategy. These observations further validate the effectiveness of our method.

We also conduct ablation studies regarding the choice of the sliding window size in the frequency domain, denoted as the window ratio $\rho$, for each model. Table~\ref{tab:ablation_dllm} presents the experimental results for LLaDA-1.5B and LLaDA-8B-Instruct under different settings. Based on the overall performance across downstream tasks, we selected 0.2 and 0.4 as the window ratios for the two models, respectively.

\section{Discussion}\label{sec:diss}
\subsection{Analysis of Decoding Block Size}\label{subsec:dissofblock}

To verify the effectiveness of FourierSampler under different decoding block sizes for dLLMs, we further conduct an experiment shown in Table~\ref{tab:blocksize_ablation} and observe that as the block size increases, applying FourierSampler to downstream tasks leads to more pronounced performance gains. This is because larger blocks provide a more complete and continuous signal for frequency-domain analysis, allowing the Fourier transform to more accurately capture low-frequency components that correspond to global semantics and structural information. When the block size is too small, the sequence is frequently segmented, which can fragment frequency-domain representations across blocks and make low-frequency information difficult to localize reliably. In contrast, larger blocks allow frame-level information to be fully distilled and well modeled at early stages, and to consistently guide subsequent fine-grained generation. As a result, the advantages of the FourierSampler become more evident in downstream task performance. It can be further observed that after applying FourierSampler, the scores on MBPP and GSM8K at block size $B=64$ are both higher than the baseline at block size 32 or 16, suggesting that our method effectively mitigates, and in some cases even avoids the severe performance degradation when block size increases. 

\subsection{Analysis of Generation Order}

\begin{figure*}[!t]
    \centering
    \hspace*{-0.4cm}%
    \includegraphics[width=1.1\linewidth]{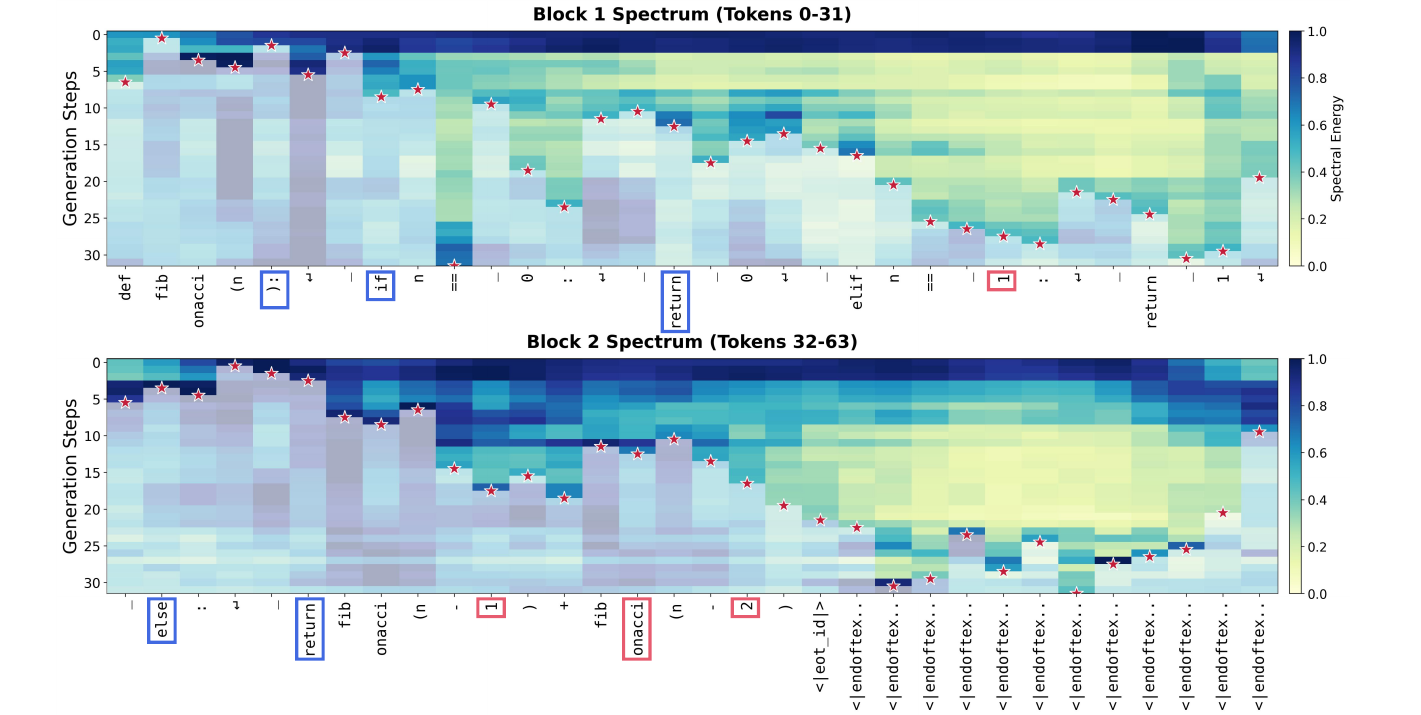}
    \caption{Visualization of step-wise generation trajectory for the prompt ``Write a python function to compute Fibonacci sequence'' on LLaDA-8B-Instruct~\citep{nie2025large}. The heatmap displays the Translated Fourier Score $\ell_{s,t}$ at each step, with red stars marking the precise step where each token is decoded. We highlight blue boxes that correspond to structure words and are decoded in the early stages, and red boxes that correspond to detail words and are filled in later stages, which validates the ``structure-to-detail'' decoding pattern of FourierSampler.}
    \label{fig:order}
\end{figure*}


To investigate whether the generation trajectory under FourierSampler aligns with the spectral characteristics observed in the static forward pass discussed in Section~\ref{subsec:freq} and truly activates the non-autoregressive potential of dLLMs, we visualize the step-by-step decoding process of LLaDA-8B-Instruct~\citep{nie2025large} on a code generation task. The heatmap visualizes, for two decoding blocks (Block~1 and Block~2), the Translated Fourier Score $\ell_{s,t}$ computed at each token position under the frequency band selected for each generation step. Red stars indicate the generation step at which each token is ultimately finalized. A structure-to-detail generation pattern can be observed in Figure~\ref{fig:order}.

From the spectrum of Block~1, it is evident that keywords representing the logical skeleton of the program—such as \textit{if} (position~9), and \textit{return} (positions~17 and~28)—achieve high scores and are determined at very early decoding stages (Steps~0--10). This indicates that the model prioritizes constructing the overall structural framework of the code. In contrast, specific variable names (e.g., \textit{fib}, \textit{n}) and numerical values (e.g., \textit{0}, \textit{1}) are generally generated at later decoding stages (Steps~15--30). For instance, in Block~2, although \textit{else} (position~33) and \textit{return} (position~37) appear relatively early, the subsequent concrete computation logic involving \textit{fib} (position~45) and \textit{n} (position~47) does not emerge until around Step~20.

This structure-to-detail generation trajectory provides intuitive evidence that our Translated Fourier Score successfully maps low-frequency energy in the frequency domain to structural information in text. As a result, dLLM can plan global logic first and subsequently fill in local details.

\subsection{Analysis of Part-of-Speech}


To investigate which words in natural language likely correspond to high-frequency components and which favor low-frequency ones, we conducted a detailed statistical analysis on the WikiText-103 dataset~\citep{merity2016pointer}. For each paragraph, we extract the final-layer hidden state sequence $\bm{H} \in \mathbb{R}^{L \times D}$, apply the spectral filtering mechanism defined in Equation~\ref{fourierfilter}, and calculate the low-frequency ratio $r_\text{low}$ as defined in Equation~\ref{r_low}. Based on this, tokens with $r_\text{low} > 0.5$ (indicating low-frequency dominance) are classified into the \textit{low group}, while others are assigned to the \textit{high group}. Figure~\ref{fig:partofspeech} illustrates the distribution ratios of different parts of speech across these two groups.

Function words and connectives, which are primarily responsible for constructing sentence logic and structural scaffolding~\citep{carnap1937logical,ru2023distributed,liu2024longwanjuan}, occupy a larger proportion of the low-frequency group. In particular, conjunctions (e.g., \textit{but}, \textit{if}, \textit{because}), prepositions (e.g., \textit{in}, \textit{for}), and adverbs (e.g., \textit{firstly}) exhibit the highest ratios of low-frequency dominance. In addition, verbs, as the core predicates of sentences, also show a strong low-frequency tendency. This observation explains why control-flow tokens such as \textit{def} and \textit{return} are preferentially generated in the code-generation heatmap presented in the previous section.

In contrast, nouns exhibit the strongest high-frequency characteristics among all part-of-speech categories. Nouns typically refer to concrete entities, variables, or values (e.g., \textit{fib}, \textit{n}, \textit{0}), serving as the specific content that fills the syntactic skeleton. These results are consistent with our distinction between \textit{framework words} and \textit{detail words} in natural language. FourierSampler leverages this property to transform implicit linguistic hierarchies into explicit generation planning.

\section{Conclusions}
In this work, we investigate the internal decoding mechanisms of dLLMs from the perspective of signal processing. We conduct the first frequency analysis in
dLLMs, showing that low-frequency implies structure, and high-frequency implies detail. Then, we propose FourierSampler. By leveraging the Translated Fourier Score and Adaptive Fourier Calibrator, our method dynamically guides the dLLMs to achieve a structure-to-detail generation.

Extensive experiments across full-bidirectional-attention (LLaDA Series) and block-wise causal attention (SDAR Series) architectures demonstrate that FourierSampler achieves consistent performance improvements in different tasks, such as math and code. Furthermore, our analyses regarding different decoding block sizes, generation order, and part-of-speech distributions further corroborate the rationality of FourierSampler. This study not only experimentally surpasses similarly sized auto-regressive models but also paves an endogenous way for future research to unlock the arbitrary-order generation potential of dLLMs.









\bibliographystyle{plainnat}
\bibliography{main}

\clearpage
\beginappendix



\section{Details of Method}\label{sec:more_details}

The key hyperparameters of our FourierSampler for different dLLMs are shown in Table~\ref{tab:hyper}.

\begin{table}[!h]
\centering
\small
\tabcolsep=0.13cm
\begin{tabular}{lcccc}
\toprule
 & $\epsilon$ & $\rho$ & $\beta_{\min}$ & $\beta_{\max}$ \\
\midrule
LLaDA1.5-8B & 1e-5 & 0.2 & 0.4 & 0.6 \\ 
LLaDA-8B-Instruct & 1e-5 & 0.4 & 0.4 & 0.6 \\ 
\bottomrule
\end{tabular}
\caption{Hyper-parameters.}
\label{tab:hyper}
\end{table}

As we have presented in Section~\ref{sec:method}, we introduce an adaptive weight $\beta_s$ based on the original decoding confidence. At the decoding step $s$, it is calculated based on the variance $\sigma_s^2$ over the maximum probability $q_{s,t}$ in the prediction distribution at each masked position $t\in\mathcal{M}_s$. We record the $\sigma_s^2$ over the past 20 decoding steps, compute the percentile $p_s$ of the current variance within the history, and linearly map it to the effective support interval $[-3,3]$ of the standard normal distribution, then obtain a smooth value $w_s\in(0,1)$ through the cumulative distribution function of the normal distribution, $F(x)=\frac{1}{2}*\left(1+\text{erf}\left(\frac{x}{\sqrt{2}}\right)\right)$. The pseudocode of the whole process is shown in Alg~\ref{alg:beta}.

\begin{algorithm}[H]
\caption{Compute Adaptive Weight $\beta_s$}\label{alg:beta}
\begin{algorithmic}[1]
\Procedure{ComputeAdaptiveWeight}{}
\If {not $\mathcal{M}_s$.any()} 
\State \Return $\beta_{\min}$
\EndIf
\State $\sigma_s^2=\text{Var}\left(\left\{q_{s,t}\right\}_{t\in\mathcal{M}_s}\right)$
\State v\_list.append$\left(\sigma^2\right)$
\If {len(v\_list) $> 20$} 
\State v\_list.pop(0)
\EndIf
\If {len(v\_list) $== 0$} 
\State $p_s = \frac{1}{2}$
\Else 
\State v\_list\_ = [\_ for \_ in v\_list if \_ $< \sigma_s^2$)]
\State $p_s = \dfrac{\text{len}\left(\text{v\_list\_}\right)}{\text{len}\left(\text{v\_list}\right)}$
\EndIf
\State $z_s$ = $\left(p_s - \frac{1}{2}\right) * 3$
\State $w_s = \frac{1}{2} * \left(1 + \text{erf}\left(\frac{z_s}{\sqrt{2}}\right)\right)$
\State $\beta_s = \beta_{\min} + (1 - w_s) * (\beta_{\max} - \beta_{\min})$
\State \Return $\beta_s$, v\_list
\EndProcedure
\end{algorithmic}
\end{algorithm}

\clearpage

\section{Details of analysis}

\begin{figure*}[!t]
    \centering
    \includegraphics[width=0.9\linewidth]{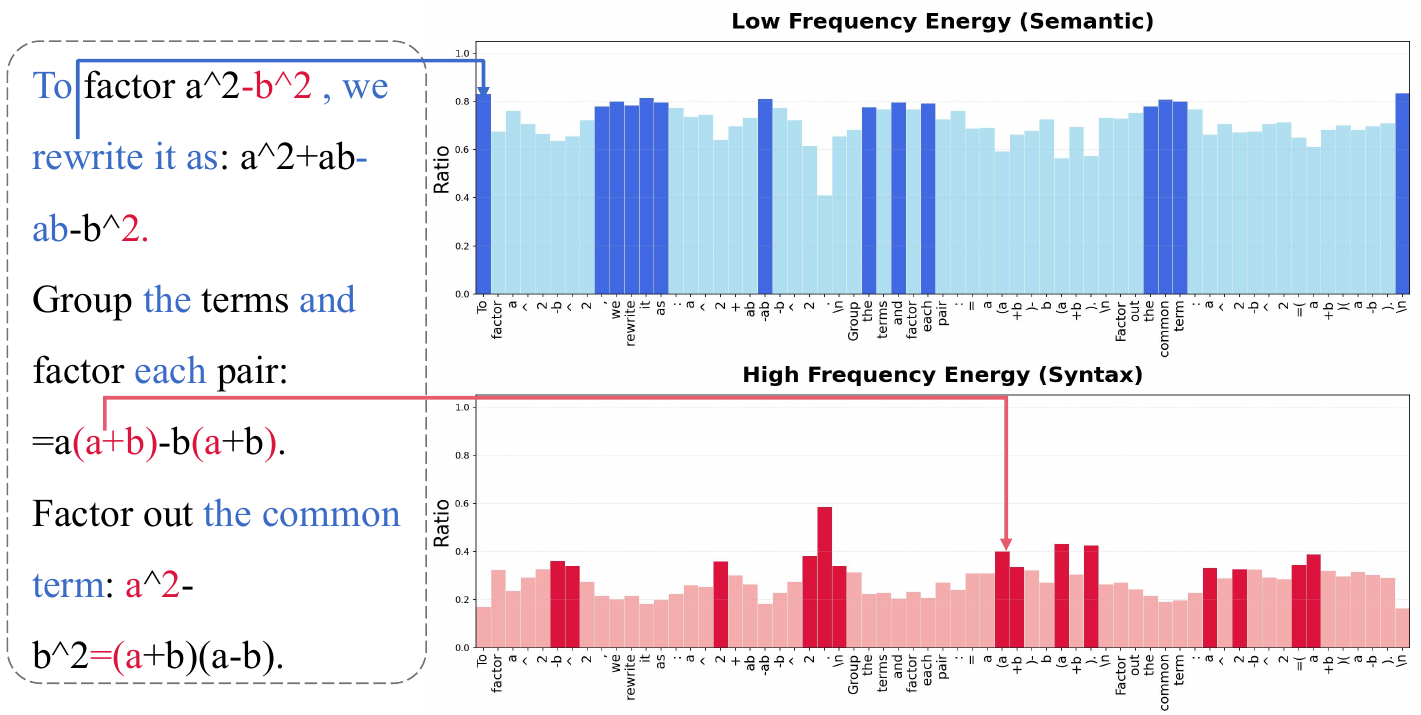}
    \caption{Visualization of the correspondence between frequency-domain analysis and textual information.}
    \label{fig:observation2}
\end{figure*}

Beyond code generation tasks, we also conducted a frequency-domain analysis on a mathematical derivation passage concerning the \textbf{difference of squares formula}. As shown in Figure \ref{fig:observation2}, narrative text appears as low-frequency components, while specific formulas and variables emerge as high frequency. This observation further corroborates our experimental findings in Section~\ref{subsec:freq}.

\end{document}